\documentclass[11pt]{article}

\usepackage[final]{acl}
\usepackage{comment}
\usepackage{tabularx}
\usepackage{amsmath}

\usepackage{times}
\usepackage{latexsym}

\usepackage[table]{xcolor}
\usepackage{array}
\usepackage{booktabs}
\usepackage{multirow}
\usepackage[most]{tcolorbox}
\usepackage{soul}

\usepackage[T1]{fontenc}

\usepackage[T1]{fontenc}

\usepackage[utf8]{inputenc}

\usepackage{microtype}
\usepackage{inconsolata}
\usepackage{graphicx}
 \usepackage{enumitem}

\title{Structured Claim-Level Discourse Representations for \\ Dense Health Narratives}

\author{
  Farnoushsadat Nilizadeh\textsuperscript{1}\thanks{These authors contributed equally to this work.} \quad 
  Elham Pourabbas Vafa\textsuperscript{2$*$} \quad 
  Shirin Nilizadeh\textsuperscript{2} \quad 
  Eduard Dragut\textsuperscript{1} \\[0.5em]
  \textsuperscript{1}Temple University \quad \textsuperscript{2}University of Texas at Arlington \\
  \texttt{\{farnoushsadat.nilizadeh, edragut\}@temple.edu}, \quad 
  \texttt{\{elham.pourabbasvafa, shirin.nilizadeh\}@uta.edu}
}

\begin{document}
\maketitle

\begin{abstract}
Health discourse in social media videos often contains densely entangled claims spanning multiple thematic aspects, stances, evidential frames, and rhetorical functions within short conversational spans. Existing approaches largely rely on coarse topic-level, sentiment-based, or stance-oriented representations that do not adequately capture this structure. Our analysis identifies an average of 13.22 atomic claims per minute, motivating richer claim-level discourse representations.
We introduce a structured framework for claim-level discourse analysis in dense health narratives. Our framework models discourse through tuples linking atomic claims with thematic aspects, stance, and multidimensional pragmatic discourse attributes. To support this setting, we construct a benchmark spanning four health domains with 1,191 manually annotated claims from 60 videos.
Using this framework, we evaluate automated structured discourse analysis under different discourse context settings. Results show that current LLMs achieve strong performance on thematic categorization and stance prediction, but struggle with high-dimensional pragmatic profiling. We also find that different discourse tasks benefit from different forms of contextual reasoning, suggesting that future systems may require task decomposition and specialized inference strategies.

\end{abstract}

\section{Introduction}
\label{Introduction}

Short-form social media videos are major sources of health information, yet health-related discourse in these environments is highly informal, rhetorically heterogeneous, and difficult to analyze computationally~\cite{duan2025crowdsourcing,fong2024ozempic,han2024public,yeung2025online,basch2023descriptive,javaid2024trends}. Existing computational approaches largely rely on coarse topic-level classification, sentiment analysis, or stance detection~\cite{augenstein2016stance,hosseinia-etal-2020-stance}, providing limited support for modeling the internal structure of dense health narratives. In practice, a single conversational segment may contain multiple atomic claims spanning different thematic aspects, conflicting stances, evidential frames, uncertainties, behavioral recommendations, and rhetorical functions. Our analysis identifies an average of 13.22 atomic claims per minute across health-related narratives, revealing a level of information density that conventional discourse representations struggle to capture. We refer to this phenomenon as \textit{multi-aspect entanglement}: the interleaving of multiple claims, aspects, and communicative intents within short conversational spans.

\begin{figure*}[t]
\centering
\footnotesize
\begin{tcolorbox}[colback=gray!5,colframe=black,title=\textbf{Input: Raw Transcript Segment (TRT)}]
\textit{``But listen, TRT isn't for everyone. You need proper blood tests to qualify. If you qualify, you might notice better mood. But done wrong, you risk high blood pressure. Joe's been on TRT since his 40s; he says it helps him stay strong."}
\end{tcolorbox}

\centering \small $\downarrow$ \textbf{Aspect-Conditioned Claim Extraction} ($V \to \{(c, a, s, \vec{\tau})\}^n$) $\downarrow$

\setlength{\tabcolsep}{6pt}
\renewcommand{\arraystretch}{1.1}

\begin{tabularx}{\textwidth}{l X >{\raggedright\arraybackslash}p{3.7cm} c l}
\hline
\# & \textbf{Atomic Claim ($c$)} & \textbf{Aspect ($a$)} & \textbf{Stance} & \textbf{Typology Vector ($\vec{\tau}$)} \\ \hline
1 & TRT isn't for everyone. & Clinical Diagnosis \& Patient Eligibility & Neu & [Non, FaC, Gen, Abs, Tml, Eva] \\ \hline
2 & You need proper blood tests to qualify for TRT. & Clinical Diagnosis \& Patient Eligibility & Neu & [Con, Pol, Gen, Abs, Tml, Des] \\ \hline
3 & Qualifying for TRT can lead to a better mood. & Mental \& Cognitive Impact & Pos & [Cau, FaC, Gen, Hed, Tml, Des] \\ \hline
4 & Taking TRT incorrectly can lead to high blood pressure. & Medical Health \& Biomarkers & Neg & [Cau, FaC, Gen, Hed, Tml, Des] \\ \hline
5 & Joe's been on TRT since his 40s. & Clinical Diagnosis \& Patient Eligibility & Neu & [Non, FaL, Ane, Att, Cur, Nar] \\ \hline
6 & Joe says TRT helps him stay strong physically. & Muscle, Performance \& Aesthetics & Pos & [Cau, FaC, Ane, Att, Cur, Nar] \\ \hline
\end{tabularx}

\vspace{1mm}
\scriptsize
\raggedright
\textbf{Typology Legend ($\vec{\tau}$):} 
\textbf{L}ogic (\textbf{Non}-logical, \textbf{Con}ditional, \textbf{Cau}sal); 
\textbf{V}erifiability (\textbf{FaC}:Factual(Clinical/Biological), \textbf{FaL}:Factual(Logistical/Contextual), \textbf{Pol}icy(Prescriptive)); 
\textbf{E}vidence (\textbf{Gen}eral Knowledge/Common Sense, \textbf{Ane}cdotal/Personal); 
\textbf{C}ertainty (\textbf{Abs}olute/Imperative, \textbf{Hed}ged(Probabilistic) , \textbf{Att}ributed/Distanced); 
\textbf{T}emporality (\textbf{Tml}: Timeless, \textbf{Cur}rent/Past); 
\textbf{F}ocus (\textbf{Eva}luative, \textbf{Des}criptive/Definitional, \textbf{Nar}rative/Sequential).
\caption{An illustration of our formal task: decomposing high-density narratives into structured tuples. This segment demonstrates \textbf{multi-aspect entanglement} and shifting rhetorical profiles, moving from prescriptive medical policy (Row 2) to hedged physiological outcomes (Rows 3--4) and attributed personal narrative (Row 6).}
\label{fig:claim_deconstruction}
\vspace{-10pt}
\end{figure*}

Prior work in fine-grained sentiment analysis, including aspect-based sentiment analysis \cite{dragut2010construction,schneider-dragut-2015-towards} and aspect-based sentiment intensity analysis \cite{dragut-fellbaum-2014-role,wang-dragut-2024-overlooked}, demonstrated the value of decomposing opinions into aspect-specific representations rather than assigning document-level labels \cite{ACOS-Cai-21ACL,RefineABSA-Su-24ACL,CompoundABSALLM-Bai-24EMNLP}. 
Dense health narratives, however, involve intertwined claims and pragmatic discourse phenomena that extend beyond sentiment-oriented representations.
We argue that dense health narratives require claim-level representations that jointly model atomic claims, thematic aspects, stance, and pragmatic discourse structure. To address this problem, we represent each narrative as a collection of $(c, a, s, \vec{\tau})$ tuples linking an atomic claim $c$ with its thematic aspect $a$, stance $s$, and multidimensional typology $\vec{\tau}$. This decomposition enables fine-grained computational analysis of dense conversational discourse.

As illustrated in Figure~\ref{fig:claim_deconstruction}, a single short-form health narrative may contain medical eligibility criteria, health outcomes, anecdotal evidence, and persuasive framing within the conversational span. Our representation separates these intertwined components into distinct claim-level structures while preserving their pragmatic and rhetorical properties. By moving beyond coarse topic- and sentiment-level labels, the proposed framework enables fine-grained modeling of dense health discourse.

To study this problem, we introduce the first benchmark for structured claim-level discourse analysis of dense health narratives. The benchmark spans four health domains: \emph{GLP-1 weight-loss medications}, \emph{Testosterone replacement therapy (TRT)}, \emph{Collagen supplementation}, and \emph{Intermittent fasting}. Constructing the benchmark required extensive expert reconciliation to address the rhetorical ambiguity, implicit framing, and conversational variability characteristic of online health discourse.

Using this benchmark, we evaluate automated structured discourse analysis under varying contextual and inference settings. Our experiments show that LLM-based systems achieve strong performance on thematic and stance classification while degrading substantially on pragmatic discourse modeling. We further find that different discourse tasks require different forms of contextual reasoning: aspect classification benefits from localized semantic context, whereas pragmatic discourse modeling relies more on sequential conversational structure. 

Our contributions are as follows:

\begin{itemize}[nosep]
   
\item We introduce the task of structured claim-level discourse analysis for dense health narratives.
    
\item We propose a multidimensional representation linking atomic claims with thematic aspects, stance, and pragmatic attributes.
    
\item We construct a benchmark spanning four health domains with 1,191 manually annotated claims from 60 videos.
    
\item We evaluate automated structured discourse analysis and study how different contextual settings affect discourse prediction tasks.

\end{itemize}

\section{Related Work}
Prior work on claim mining and argument analysis has progressively evolved from surface-level claim detection toward richer structured representations \cite{Fanetal20,habernal2017argumentation,lawrence2019argument,lippi2015context,lippi2016argument}. More recent approaches jointly model claims with stance, argumentative structure, semantic aspects, and rhetorical context rather than treating claims as isolated propositions \cite{trautmann2020fine,hosseinia-etal-2021-usefulness,jo2020machine,plenz2025argumentation,guo2024disordered,guo2024modabs}. Parallel work in discourse analysis and evidentiality modeling has further introduced pragmatic dimensions such as certainty, evidential grounding, communicative intent, and logical structure into discourse representations \cite{daxenberger2017essence,boland2022beyond,schaefer2023towards,dhole2025conqret,visser2020annotating}.

Collectively, these works highlight the importance of modeling semantic, argumentative, and pragmatic structure jointly. However, existing benchmarks typically study these dimensions in isolation or within comparatively constrained discourse settings. Our work instead focuses on dense health narratives where multiple intertwined claims, aspects, evidential strategies, and rhetorical intents coexist within short conversational spans. 
Appendix~\ref{sec:relatedwork-appendix}  gives additional related discussion.

\section{Structured Discourse Representation}
\label{sec:representation}
In this section, we introduce a claim-level representation framework for dense discourse environments that jointly models atomic claims, thematic aspects, stance, and pragmatic discourse structure. Structured representations play a central role in NLP, including semantic role labeling, frame semantics, discourse parsing, argument mining, and aspect-based sentiment analysis~\cite{lippi2016argument,augenstein2016stance}. Existing formulations, however, provide limited support for jointly modeling multiple semantic and pragmatic dimensions at the claim level. We therefore represent dense discourse through structured tuples:
\[
\mathcal{T}_V = \{(c,a,s,\vec{\phi})_i\}_{i=1}^{n}, \text{where}
\]

\begin{itemize}[nosep]
    \item $c$ denotes an atomic claim,
    
    \item $a \in \mathcal{A}$ denotes the thematic aspect associated with the claim,
    
    \item $s \in \{+, -, 0\}$ denotes the claim stance,
    
    \item $\vec{\phi} = (\phi_1,\ldots,\phi_6)$ denotes an ordered pragmatic discourse profile, where each component $\phi_j \in \mathcal{D}_j$ is drawn from a dimension-specific categorical domain.
\end{itemize}

The proposed framework consists of two complementary semantic layers. The first captures \textit{domain-dependent thematic structure}, represented through aspects and stance. The second captures \textit{domain-stable pragmatic discourse structure}, represented through rhetorical and communicative dimensions. This distinction is intuitive in dense discourse environments. For example, Ozempic-related narratives involve aspects such as appetite suppression, gastrointestinal side effects, insurance access, and weight-loss effectiveness, whereas Testosterone Replacement Therapy (TRT) narratives emphasize hormonal regulation, fertility, libido, and physical performance. In contrast, rhetorical structures such as anecdotal evidence, causal reasoning, speculative framing, uncertainty, and persuasive recommendations recur consistently across both domains.

The proposed framework is a flexible claim-level representation rather than a fixed schema. Though developed for health-related social media discourse, the framework may generalize to other domains involving complex narrative structure, including financial advice, political commentary, scientific debate, and legal discourse. We leave broader cross-domain evaluation to future work.

\subsection{Atomic Claim Decomposition}

Given a narrative or transcript segment $V$, our first objective is to decompose the discourse into a set of atomic claims
$
V \rightarrow \{c_i\}_{i=1}^{n},
$
where each $c_i$ denotes a self-contained propositional unit expressing a single coherent assertion. This decomposition separates compound conversational structures into individually analyzable claim units. For example, the statement:
\emph{``Ozempic helped me lose weight, but it made me constantly nauseous,''}
contains at least two distinct claims involving different aspects and opposing stances. Treating such discourse as a single unit obscures the distinction between therapeutic benefits and adverse effects.

To address this problem, we decompose complex statements into minimal propositional units while preserving their semantic interpretation. We resolve ambiguous references following principles of anaphora grounding~\cite{lee2017end}, replacing pronouns and context-dependent expressions with explicit referents when necessary. The resulting atomic claims form the basic units for downstream analysis.

\subsection{Thematic Aspect Modeling}

Each atomic claim is associated with a thematic aspect
$a \in \mathcal{A}$ representing the primary semantic category addressed by the claim. Aspect assignments operate at the claim level, allowing different propositions within the same narrative segment to be mapped to distinct semantic categories. 
The aspect inventory $\mathcal{A}$ is inherently domain-dependent and reflects the semantic structure of a particular narrative environment. Consequently, aspects are modeled as flexible semantic categories rather than fixed universal labels.

Our formulation is conceptually related to aspect-based sentiment analysis, where opinions are decomposed according to thematic targets rather than treated as document-level judgments \cite{wang2026danceha}. Dense narrative settings, however, require modeling substantially richer semantic and pragmatic structure. 

\subsection{Pragmatic Discourse Instantiation}

While aspects capture thematic content, they provide limited support for modeling rhetorical and pragmatic discourse structure. To address this limitation, we associate each claim with a multidimensional typology vector:
\[
\vec{\phi} = [l_1,l_2,\dots,l_6]
\]

where each dimension characterizes a distinct discourse property.

The proposed typology is synthesized from prior work in discourse analysis, argumentation theory, evidentiality, stance modeling, uncertainty detection, and pragmatic classification~\cite{lippi2016argument,augenstein2016stance}.  The difficulty of defining stable fine-grained intent taxonomies has also
been observed in other structured language-analysis settings
~\cite{lan-etal-2025-unit}. During development, we evaluated a broader inventory of candidate dimensions and retained only those that consistently appeared across multiple discourse domains and could be reliably annotated under expert reconciliation.
This selection reflects a balance between expressiveness and annotatability rather than a claim that the six dimensions are exhaustive of all possible pragmatic phenomena in health discourse. The resulting framework should be viewed as an extensible discourse representation rather than a closed label schema.

Table~\ref{tab:typology_summary} summarizes the proposed typology, with detailed  definitions provided in Table~\ref{tab:6_axis_taxonomy} in Appendix~\ref{app:6_axis_taxonomy_sec}.
 The typology models six complementary dimensions:
\textbf{Logical Relationship:} the structural relationship expressed by the claim, e.g., causal, conditional, or comparative reasoning. 
    \textbf{Verifiability and Intent:} whether the claim expresses factual, policy-oriented, subjective, or speculative content.    
    \textbf{Evidence Basis:} the implied source of authority underlying the claim, e.g., anecdotal experience, scientific evidence, expert opinion, or general knowledge.  
    \textbf{Certainty:} the degree of confidence or hedging expressed by the speaker.  
    \textbf{Temporality:} the temporal orientation of the claim, e.g., timeless principle, current experience, or future prediction.   
    \textbf{Claim Focus:} the primary communicative role of the statement, e.g., explanatory, descriptive, evaluative, or narrative.
The proposed dimensions enable fine-grained modeling of rhetorical and pragmatic structure at the claim level.

\begin{table}[t]

\centering

\small

\setlength{\tabcolsep}{4pt} 

\renewcommand{\arraystretch}{1.1}

\caption{Six-dimensional pragmatic discourse typology.}

\label{tab:typology_summary}

\begin{tabularx}{\columnwidth}{l X} 

\hline

\textbf{Dimension} & \textbf{Representative Labels} \\ \hline

Logical Relationship & Causal, Comparative, Conditional \\

Verifiability \& Intent & Factual, Subjective, Prescriptive \\

Evidence Basis & Scientific, Anecdotal, Commercial \\

Certainty & Absolute, Hedged, Attributed \\

Temporality & Past, Predictive, Timeless \\

Claim Focus & Explanatory, Narrative, Promotional \\ \hline
\vspace{-10pt}
\end{tabularx}

\end{table}

For example, in Figure~\ref{fig:claim_deconstruction}, the statement ``Joe says TRT helps him stay strong physically'' is represented as an atomic claim with thematic aspect \textit{Muscle, Performance \& Aesthetics}, positive stance, and a pragmatic profile consisting of causal reasoning, factual(clinical) verifiability, anecdotal evidence, attributed certainty, past temporality, and narrative focus. In tuple form:
\[
\small
\begin{aligned}
(c, a, s, \vec{\phi}) = (&
\texttt{TRT helps him stay strong physically}, \\
& \texttt{Muscle/Performance}, \texttt{Positive}, \\
& (\texttt{Causal}, \texttt{Factual(Clinical)}, \texttt{Anecdotal},  \\
&  \texttt{Attributed},  \texttt{Past}, \texttt{Narrative}))
\end{aligned}
\]

To illustrate why simpler structures such as aspect-plus-sentiment or stance-only labels are insufficient for dense health narratives, consider the paired claims from the TRT domain in Table~\ref{tab:pragmatic_divergence}. While both share the identical thematic aspect (Fertility \& HPTA) and the same negative stance (Neg), they diverge fundamentally across every pragmatic axis: 
A traditional aspect-plus-sentiment or stance-only representation completely collapses these critical rhetorical variations, treating both assertions as identical. Our 6-axis typology is what preserves these vital distinctions, which are necessary for downstream tasks like misinformation detection and evidence quality assessment.

\begin{table*}[t]
\centering
\caption{Qualitative comparison of two atomic claims from the TRT domain sharing identical thematic aspect and stance labels yet diverging across all six pragmatic discourse dimensions, illustrating the limitation of coarse aspect-plus-sentiment or stance-only baselines.}
\footnotesize
\setlength{\tabcolsep}{3pt}
\begin{tabular}{p{3.9cm} p{1.6cm} p{0.7cm} p{1.0cm} p{1.5cm} p{1.5cm} p{1.2cm} p{1.1cm} p{1.0cm}}
\toprule
\textbf{Claim} & \textbf{Aspect} & \textbf{Stance} & \textbf{Logical} & \textbf{Verif.} & \textbf{Evidence} & \textbf{Certainty} & \textbf{Temp.} & \textbf{Focus} \\
\midrule
On average, testicular function drops approximately 50--75\%. & Fertility \& HPTA & Neg & Causal & Factual (Clin.) & Sci. (Stat.) & Absolute & Timeless & Descriptive \\
\addlinespace[0.4em]
The speaker acknowledges that 50--75\% decrease is dramatic. & Fertility \& HPTA & Neg & Non\_Log & Value (Subj.) & Anecdotal & Attributed & Current / Past & Evaluative \\
\bottomrule
\end{tabular}

\label{tab:pragmatic_divergence}
\end{table*}

A broader qualitative analysis of multi-aspect entanglement and intra-speaker rhetorical trajectories across domains is provided in Appendix~\ref{app:pragmatic_divergence}.
\section{Benchmark Construction}
\label{sec:benchmark}
We now describe the benchmark construction process for the proposed framework.

\subsection{Domain Selection and Video Collection}

We selected four health domains: \emph{GLP-1 weight-loss medications}, \emph{Testosterone Replacement Therapy (TRT)}, \emph{Collagen supplementation}, and \emph{Intermittent fasting}. These domains capture diverse forms of online health discourse, including personal narratives, lifestyle advice, clinical explanations, commercial promotion, and policy-oriented discussion.

Videos were collected using the YouTube Data API v3~\cite{youtubeapi}. All searches were constrained to a fixed collection window spanning January 1, 2024 through August 30, 2025. Initial seed queries, e.g., ``Ozempic journey'' and ``TRT side effects,'' were manually audited for relevance and expanded into broader topic-specific query sets. Using these queries, we collected an initial pool of approximately 2,000 unique video IDs per topic.

\subsection{Corpus Filtering and Sampling}

To move from the high-recall pool to a high-precision evaluation set, we implemented a multi-stage selection process. First, we extracted a randomized subset of 100 candidate videos per topic to establish a manageable working pool for intensive processing. These candidates were processed through the multimedia pipeline described in Section~\ref{metadata_harvesting} to generate denoised transcripts.
Next, an automated triage layer was applied to the transcripts using an LLM-based screening prompt (Appendix~\ref{app:triage_prompt}). This stage filtered videos according to three criteria: (i) English-language content, (ii) presence of meaningful propositional speech, and (iii) thematic relevance to the target health domain. The filtering stage reduced the candidate pool to approximately 50 high-quality videos per topic.
From this pool, we performed a final purposive selection of 15 videos per topic ($N=60$ total). The selection process additionally prioritized shorter, claim-dense videos to maximize rhetorical diversity and avoid over-representation of a single discourse type. Across the 60 selected videos, individual durations range from 15 seconds to 6 minutes 50 seconds (mean = 90.1 s, SD = 83.7 s), reflecting a corpus of short-to-medium-length content. Consequently, the final corpus includes a rich cross-section of informational, promotional, prescriptive, and narrative categories. 
The distribution of these discourse types across the 60 videos is presented in Table~\ref{tab:corpus_topic}. 

\begin{table}[t]
\centering
\small
\setlength{\tabcolsep}{3pt} 
\caption{Distribution of video discourse categories across the curated Ground Truth datasets ($N=60$).}
\label{tab:corpus_topic}
\newcolumntype{C}{>{\centering\arraybackslash}X} 
\begin{tabularx}{\linewidth}{lCCCC}
\hline
\textbf{Topic} & \textbf{Informa- tional} & \textbf{Promo- tional} & \textbf{Prescrip- tive} & \textbf{Narrative} \\ \hline
Ozempic  & 4 & 1 & 4 & 6 \\
TRT      & 7 & 3 & 3 & 2 \\
Collagen & 5 & 7 & 2 & 1 \\
Fasting  & 5 & 1 & 7 & 2 \\ \hline
\textbf{Total} & \textbf{21} & \textbf{12} & \textbf{16} & \textbf{11} \\ \hline
\end{tabularx}
\vspace{-10pt}
\end{table}

\subsection{Transcript Processing}
\label{metadata_harvesting}

Our multimedia pipeline processed the videos through a series of automated transcription and stabilization layers. Audio streams were extracted using \texttt{yt-dlp}~\cite{ytdlp} and transcribed using OpenAI's Whisper model~\cite{radford2023robust}.
\texttt{yt-dlp} handles network retrieval and audio extraction from the YouTube platform, while Whisper performs the speech-to-text transcription on the resulting audio stream.

The raw automatic speech recognition (ASR) output frequently contained malformed medical terminology, fragmented sentence boundaries, and missing punctuation (Appendix~\ref{app:denoising_examples}). To address these issues, we introduced an \emph{Orthographic Denoising and Stabilization} layer using Gemini 2.5 Flash \cite{comanici2025gemini} (Appendix~\ref{app:denoising_prompt_template}) as a post-processing step over the raw Whisper transcripts. The denoising stage performed two complementary corrections: (i) spelling normalization for ASR-induced errors in drug names, biomarkers, and domain-specific medical terminology, and (ii) orthographic stabilization correcting fragmented sentence boundaries, capitalization, and missing punctuation. The prompt enforced conservative editing constraints that prohibited content deletion, summarization, or insertion, preserving the speaker's original lexical content and propositional meaning. Examples in Appendix~\ref{app:denoising_examples} illustrate consistent recovery of degraded terminology and sentence structure.

\subsection{Gold Standard Annotation}
To construct the gold-standard benchmark, two annotators with backgrounds in Natural Language Processing (NLP) and computational discourse analysis carried out a multi-stage annotation protocol over the evaluation set. The resulting benchmark required around 400 person-hours of annotation and reconciliation effort (Table~\ref{tab:annotation_effort} in Appendix~\ref{app:annot}).

The annotation process consists of four stages. First, annotators identify discourse segments relevant to the target health domain. Second, extracted statements are decomposed into atomic claims using the principles described in Section~\ref{sec:representation}. Third, annotators assign thematic aspects and stance labels to each claim. Finally, each claim is annotated across the six-dimensional pragmatic typology introduced in Section~\ref{sec:representation}. The process requires substantial reconciliation due to implicit causality, anecdotal framing, speculative language, and overlapping communicative intent.

Table~\ref{tab:corpus_stats} summarizes descriptive statistics of the resulting benchmark. Across the four domains, the corpus contains 1,191 atomic claims extracted from approximately 90 minutes of discourse, corresponding to an average density of 13.22 atomic claims per minute. We further observe substantial variation in claim density across domains. While narrative-oriented Ozempic videos show lower density, content in the TRT, Fasting, and Collagen domains is highly condensed, peaking at 17.24.

\begin{table}[t]
\centering
\small
\setlength{\tabcolsep}{2pt}
\renewcommand{\arraystretch}{1.2}
\caption{Descriptive statistics of the curated Ground Truth (GT) datasets ($N=60$).}
\label{tab:corpus_stats}
\begin{tabular}{lccccc}
\hline
\textbf{Topic} & \textbf{Videos} & \textbf{Duration} & \textbf{Stmts} & \textbf{Claims} & \textbf{Density$^*$} \\ \hline
Ozempic  & 15 & 35m 54s & 166 & 292 & 8.13 \\
TRT      & 15 & 19m 12s & 185 & 331 &  17.24\\
Collagen & 15 & 15m 28s & 126 & 237 & 15.32 \\
Fasting  & 15 & 19m 30s & 175 & 331 & 16.97 \\ \hline
\textbf{Total} & \textbf{60} & \textbf{90m 4s} & \textbf{652} & \textbf{1,191} & \textbf{13.22} \\ \hline
\multicolumn{6}{l}{\scriptsize $^*$Density measured as Atomic Claims per minute of video content.}
\end{tabular}
\vspace{-10pt}
\end{table}

\subsection{Annotation Reliability}
\label{reliability_validation}

To validate benchmark robustness, we compute reliability metrics across all annotation layers. Because open-ended segment extraction lacks clearly defined true negatives, we evaluate agreement for the initial statement extraction and atomic decomposition stages using pairwise $F_1$-scores and macro-averaged agreement measures rather than chance-corrected coefficients, following prior work on argument mining and span-based discourse annotation ~\cite{mochales2011argumentation,habernal2017argumentation,artstein2008survey}.

For statement extraction, annotators achieved a mean pairwise $F_1$ of 0.93 across the 60-video corpus. For atomic decomposition, the overall macro-averaged agreement reached 92\%, indicating strong consistency in identifying claim boundaries despite the density and rhetorical complexity of the discourse. 
For the multi-axis classification tasks, inter-annotator agreement was evaluated using both Cohen's $\kappa$ and raw percentage agreement
~\cite{cohen1960coefficient,artstein2008survey}. As summarized in Table~\ref{tab:iaa_summary} (see Table~\ref{tab:iaa_master} for details), aspect assignment achieved a mean raw agreement of 85\% (mean $\kappa = 0.83$), reflecting the stability of the induced aspect taxonomies across domains.

\begin{table}[t]
\centering
\small
\setlength{\tabcolsep}{4pt}
\renewcommand{\arraystretch}{1.1}
\caption{Inter-annotator agreement across layers.}
\label{tab:iaa_summary}
\begin{tabular}{lc}
\hline
\textbf{Annotation Layer} & \textbf{Mean $\kappa$ / Agreement} \\ \hline
Statement Extraction & Pairwise $F_1 = 0.93$ \\
Atomic Decomposition & Agreement = 0.92 \\
Aspect Assignment & $\kappa = 0.83$ /Agreement = 0.85 \\
Stance Classification & $\kappa = 0.80$ /Agreement = 0.87 \\
6-Axis Typology & $\kappa = 0.76 $ /Agreement = 0.88   \\ \hline
\end{tabular}
\vspace{-10pt}
\end{table}

Agreement across the six-dimensional typology remained consistently substantial. Logical Relationship achieved the highest structural alignment (Mean raw agreement of 90.4\%; Mean $\kappa = 0.85$), while Certainty reached 92.9\% raw agreement. The comparatively lower agreement observed for Claim Focus (Mean $\kappa = 0.68$) reflects the inherent difficulty of distinguishing explanatory, descriptive, and narrative intent in spontaneous conversational discourse.
All remaining annotation conflicts were resolved through structured reconciliation sessions following standard adjudication practices in discourse annotation ~\cite{artstein2008survey}.

\section{Structured Discourse Prediction}
\label{sec:automation}

We now investigate whether large language models (LLMs) can  recover structured discourse tuples
$
(c,a,s,\vec{\phi})
$
from dense health narratives.

\subsection{Task Formulation}

Given an atomic claim $c$ and a contextual setting $\Gamma$, the objective is to predict its thematic aspect $a$, stance $s$, and pragmatic discourse profile $\vec{\phi}$:
\[
f_\theta(c,\Gamma) \rightarrow (a,s,\vec{\phi}),
\]

where $f_\theta$ denotes an LLM-based structured discourse prediction function.

We investigate three research questions:

\begin{itemize}[nosep]
    \item[] \textbf{RQ1:} Can LLMs reliably recover structured discourse tuples from dense health narratives?
    
    \item[] \textbf{RQ2:} How do different forms of discourse context affect discourse prediction tasks?
    
    \item[] \textbf{RQ3:} Which discourse dimensions generalize most consistently across domains?
\end{itemize}

We evaluated six inference settings that varied the amount and structure of surrounding discourse available to the model: \textbf{Atomic Claim}, where the model receives only the isolated target claim with no surrounding context; \textbf{Batch}, where claims are processed in fixed-size chronological windows of eight claims per API call, following transcript order (these windows reflect narrative co-occurrence rather than semantic retrieval); \textbf{Local ±2}, where only the two claims immediately preceding and following the target claim are provided; \textbf{Narrative}, where a higher-level narrative-arc summary of the surrounding discourse is provided; \textbf{Transcript}, where the full video transcript is provided as context; and \textbf{Summary}, where a compressed summary of the transcript is provided in place of the raw text. All results in Sections~\ref{asp_stn} and~\ref{typology_cls} use Gemini 2.5 Flash unless noted.

Our central hypothesis is that different discourse tasks require different forms of contextual reasoning. In particular, aspect assignment depends on localized semantic contrasts, whereas pragmatic discourse interpretation relies on broader conversational structure.

\subsection{Aspect Taxonomy Instantiation}
\label{aspect-discovery}

Because thematic aspects are  domain-dependent, we adopt an inductive aspect construction strategy rather than imposing a fixed predefined schema. Following prior work on iterative discourse schema development and aspect induction~\cite{jo2020machine,guo2024modabs,plenz2025argumentation}, 
we analyze representative pilot transcripts from each topic domain and use LLM-assisted claim decomposition on these transcripts to surface recurring claim targets within the domain, without constraining the output to a fixed label set. Candidate categories arising from this process are then refined through expert reconciliation, in which annotators merge, discard, or sharpen category boundaries to arrive at a finalized taxonomy for each domain. 
The induced taxonomies vary substantially across domains. For example, Ozempic discourse emphasizes appetite suppression and gastrointestinal side effects, whereas TRT discussions focus more on hormonal regulation, fertility, and physical performance. To support consistency across videos, aspect categories are defined at the domain level rather than tailored to individual narratives.

The finalized taxonomies are further validated using boundary cases containing semantically ambiguous or overlapping claims. Across the 1,191 annotated claims, only five could not be mapped cleanly to an existing aspect category, indicating broad coverage of the induced semantic categories.  Appendix ~\ref{app:aspect_taxonomies} gives the aspect taxonomies for all four domains in Table \ref{tab:aspect_taxonomies_table}.

\subsection{Automated Labeling Setup}

To scale annotation beyond the manually curated benchmark, we use an automated labeling framework based on Gemini 2.5 Flash. All generation and evaluation pipelines use deterministic greedy decoding ($\tau = 0.0$) and structured JSON outputs. 
The prompt templates for both aspect/stance classification and pragmatic typology prediction are in Appendix~\ref{app:labeling_prompt}; domain-invariant axis definitions follow Table~\ref{tab:6_axis_taxonomy}, while domain-specific aspect taxonomies follow Table~\ref{tab:aspect_taxonomies_table}. 
We use two separate prompts for these tasks. The first prompt handles joint aspect and stance prediction, receiving the domain-specific aspect schema together with category definitions and representative keywords. The second prompt handles the six-dimensional pragmatic typology.
To improve robustness under ambiguous discourse conditions, both prompts require a brief intermediate justification before final predictions. For pragmatic discourse classification, the model additionally receives surrounding transcript context to help resolve ambiguity across pragmatic dimensions. 

\begin{table}[h]
\centering
\caption{Core pipeline constants.}
\small
\begin{tabular}{|l|l|}
\hline
\textbf{Constant} & \textbf{Value} \\
\hline
Backbone model (primary) & Gemini 2.5 Flash \\
Decoding strategy & Greedy \\
Temperature (primary runs) & $\tau = 0.0$ \\
Output format & Structured JSON \\
Batch window size & 8 claims/call \\
Local context window & $\pm$2 claims \\
Few-shot examples/domain & 5 \\
Robustness temperatures & $\tau \in \{0.0,0.2,0.5,0.8\}$ \\
Robustness trials/setting & 3 (12 runs total) \\
\hline
\end{tabular}

\label{tab:hyperparams}
\end{table}

Our objective is not to optimize specialized discourse architectures, but to establish a robust benchmark and evaluation setting for structured discourse analysis. More advanced inference frameworks, including agentic and multi-stage reasoning systems, remain important directions for future work.
Table ~\ref{tab:hyperparams} summarizes the core pipeline constants used across all primary evaluations.
\section{Experiments}

We evaluate the framework in three settings: (1) aspect and stance classification, (2) multidimensional pragmatic profiling using the 6-axis typology, and (3) cross-model generalization across proprietary and open-weight LLMs. All evaluations utilize deterministic greedy decoding ($\tau = 0.0$) with fixed prompts and structured JSON outputs.

\subsection{Aspect and Stance Classification}
\label{asp_stn}
We first evaluate thematic aspect assignment and stance classification under different contextual settings. We benchmarked five contextual configurations ranging from isolated atomic claims to full transcripts, where the Batch setting processes fixed-size chronological windows of 8 claims per API call based on narrative order, and the Local context setting provides the two immediately neighboring claims around a single target claim. Results are summarized in Table~\ref{tab:ablation_results}. Batch-based inference achieved the strongest overall performance, outperforming both isolated claim classification and broader contextual settings. In contrast, full-transcript inference consistently reduced performance, suggesting that excessive conversational context introduces semantic noise.

We further evaluated prompt refinement strategies using the batch configuration. Adding category definitions and representative keywords substantially improved aspect assignment performance, while lightweight chain-of-thought prompting produced additional gains for stance classification. The final configuration (Exp. Batch + CoT) achieved 79.23\% aspect accuracy ($\kappa = 0.77$) and 91.55\% stance accuracy ($\kappa = 0.87$) on the Ozempic dataset. 
These findings suggest that thematic discourse tasks benefit from localized semantic context and lightweight reasoning support, whereas excessive discourse context may dilute the model's focus on the target claim.

\begin{table}[t]
\centering
\footnotesize
\setlength{\tabcolsep}{3.5pt}
\renewcommand{\arraystretch}{1.12}
\caption{Context and prompting ablations for aspect and stance classification using Gemini 2.5 Flash. Each cell reports Acc. / $\kappa$.}
\label{tab:ablation_results}
\begin{tabular}{llcc}
\toprule
  \textbf{Configuration of Experiment} & \textbf{Aspect} & \textbf{Stance} \\
\midrule
 Atomic Claim      & 69.71 / .66 & 85.04 / .77 \\
 \textbf{Batch} 
  & \textbf{73.72 / .70} & \textbf{90.51 / .85} \\
 Local ($\pm 2$)   & 68.98 / .65 & 85.04 / .76 \\
 Narrative         & 71.90 / .68 & 85.04 / .76 \\
 Transcript        & 69.34 / .65 & 83.58 / .74 \\
\midrule
 Batch + Defs. \& Keywords 
   & 79.16 / .77 & 89.05 / .83 \\
\textbf{Batch + CoT} 
   & \textbf{79.23 / .77} & \textbf{91.55 / .87} \\
\bottomrule
\end{tabular}
\vspace{-10pt}
\end{table}

\subsection{Pragmatic Typology Classification}
\label{typology_cls}
Next, we evaluate automated classification of the six-dimensional pragmatic discourse typology . Unlike thematic aspect labeling, pragmatic profiling requires modeling rhetorical and structural properties such as evidential framing, temporality, and communicative intent.

\begin{table*}[t]
\centering
\small
\setlength{\tabcolsep}{3.0pt}
\renewcommand{\arraystretch}{1.2}
\caption{Impact of discourse context and chain-of-thought (CoT) prompting on 6-axis typology classification using Gemini 2.5 Flash. The left portion of the table evaluates different contextual settings ranging from isolated atomic claims to broader conversational contexts. The right portion reports the effect of heuristic prompting and explicit CoT reasoning on the optimized pragmatic discourse classifier. Each cell reports Accuracy and Cohen's $\kappa$.}
\label{tab:6axis_combined}
\begin{tabular}{lcccccccccccc|cccc}
\toprule
 & \multicolumn{2}{c}{\textbf{Atomic}} 
 & \multicolumn{2}{c}{\textbf{Batch}} 
 & \multicolumn{2}{c}{\textbf{Local $\pm 2$}} 
 & \multicolumn{2}{c}{\textbf{Narr.}} 
 & \multicolumn{2}{c}{\textbf{Trans.}} 
 & \multicolumn{2}{c|}{\textbf{Summ.}}
 & \multicolumn{2}{c}{\textbf{Heuristic}} 
 & \multicolumn{2}{c}{\textbf{+ CoT}} \\
\cmidrule(lr){2-3}\cmidrule(lr){4-5}\cmidrule(lr){6-7}
\cmidrule(lr){8-9}\cmidrule(lr){10-11}\cmidrule(lr){12-13}
\cmidrule(lr){14-15}\cmidrule(lr){16-17}
\textbf{Axis} 
& Acc. & $\kappa$ 
& Acc. & $\kappa$ 
& Acc. & $\kappa$ 
& Acc. & $\kappa$ 
& Acc. & $\kappa$ 
& Acc. & $\kappa$
& Acc. & $\kappa$
& Acc. & $\kappa$ \\
\midrule
I. Logical Rel.   
& 55.91 & 0.31 
& \textbf{57.71} & \textbf{0.38} 
& 55.56 & 0.34 
& 54.84 & 0.33 
& 57.71 & 0.33 
& 57.35 & 0.33
& \textbf{84.23} & \textbf{0.74}
& 80.29 & 0.68 \\

II. Verifiability 
& 82.80 & 0.69 
& 81.72 & 0.68 
& 84.59 & 0.73
& \textbf{84.95} & \textbf{0.74} 
& 81.00 & 0.66 
& 82.08 & 0.69
& \textbf{86.38} & \textbf{0.76}
& 82.08 & 0.69 \\

III. Evidence     
& \textbf{79.57} & \textbf{0.68} 
& 62.37 & 0.50 
& 77.78 & 0.65 
& 58.78 & 0.46 
& 44.44 & 0.31 
& 53.76 & 0.40
& \textbf{84.23} & \textbf{0.74}
& 61.65 & 0.48 \\

IV. Certainty     
& \textbf{93.91} & \textbf{0.83} 
& 92.47 & 0.80 
& 91.40 & 0.77 
& 90.32 & 0.75 
& 86.38 & 0.66 
& 92.11 & 0.79
& 90.32 & 0.75
& \textbf{91.04} & \textbf{0.76} \\

V. Temporality    
& \textbf{71.33} & \textbf{0.52} 
& 63.80 & 0.42 
& 62.72 & 0.41 
& 66.67 & 0.46 
& 60.93 & 0.37 
& 68.82 & 0.49
& \textbf{91.40} & \textbf{0.84}
& 74.19 & 0.57 \\

VI. Claim Focus   
& 52.69 & 0.35 
& 53.05 & \textbf{0.38}
& \textbf{53.76} & 0.37 
& 51.61 & 0.36 
& 46.24 & 0.26 
& 41.22 & 0.26
& \textbf{82.08} & \textbf{0.74}
& 49.46 & 0.34 \\
\midrule
\textbf{Mean} 
& \textbf{72.70} & \textbf{0.57} 
& \textbf{68.52} & \textbf{0.53} 
& \textbf{70.97} & \textbf{0.54} 
& \textbf{67.86} & \textbf{0.52} 
& \textbf{62.78} & \textbf{0.43} 
& \textbf{65.89} & \textbf{0.49}
& \textbf{86.44} & \textbf{0.76}
& \textbf{73.12} & \textbf{0.59} \\
\bottomrule
\end{tabular}
\vspace{-10pt}
\end{table*}

We evaluate multiple discourse context settings ranging from isolated claims to full transcripts and compressed summaries. Results are shown on the left side of Table~\ref{tab:6axis_combined}. In contrast to thematic aspect assignment, broader contextual windows do not consistently improve performance. Full-transcript and narrative-level contexts reduce performance, particularly for structurally sensitive dimensions such as Evidence Basis and Claim Focus. Instead, the isolated claim setting (Atomic) emerged as the strongest baseline context ($72.70\%$, $\kappa = 0.57$).

We refined prompting using heuristic discourse constraints  from annotation reconciliation sessions acting as a proof of concept for rule-augmented pragmatic classification. Under this configuration, the local-context setting paired with these heuristics achieved the strongest overall performance, reaching 86.44\% mean accuracy ($\kappa = 0.76$).

Finally, we evaluated explicit chain-of-thought (CoT) prompting with the optimized typology classifier, as shown in the right-hand portion of Table~\ref{tab:6axis_combined}. Unlike aspect classification, CoT prompting substantially degraded performance across most pragmatic dimensions, particularly Claim Focus and Evidence Basis, dropping the mean accuracy to 73.12\% ( see Appendix~\ref{sec:cot_error_appendix} for a qualitative error analysis of these CoT failure modes).  Table~\ref{tab:6axis_phase2_mix} in Appendix~\ref{app:phase2_heuristics} outlines the full heuristic optimization across all other context windows.

These results show a distinction between semantic categorization and pragmatic discourse modeling. While lightweight reasoning improved thematic classification, explicit CoT prompting harmed high-dimensional pragmatic profiling. This suggests that pragmatic interpretation relies more  on implicit rhetorical cues and localized conversational structure than on explicit reasoning chains.

To evaluate stability, we reran the TRT pipeline across temperatures $\tau \in \{0.0, 0.2, 0.5, 0.8\}$ (3 trials each), demonstrating robust empirical consistency across runs alongside minor, task-specific temperature sensitivities
(Appendix~\ref{sec:robustness_appendix}).

\subsection{Cross-Model Generalization}

To evaluate the robustness of the framework across backbone models, we instantiated the identical inference pipeline using three additional models: Llama 3.3 \cite{grattafiori2024llama}, Qwen 3.5 35B \cite{qwen3.5}, and Gpt-oss 20B \cite{agarwal2025gpt}. All models were evaluated under identical prompting and contextual settings.

Table~\ref{tab:cross_model} displays Gemini 2.5 Flash and the top open-source follower, Qwen 3.5 35B; four-model results are in Appendix~\ref{sec:cross_model_full} (Table~\ref{tab:cross_model_full})
Across all domains, Gemini 2.5 Flash achieved the strongest overall performance, particularly on the 6-axis typology task. However, open-weight models remained competitive across aspect and stance classification, with Qwen 3.5 35B consistently approaching proprietary-model performance.

The largest performance divergence emerged in the pragmatic typology task, where smaller open-weight models degraded substantially under the high-dimensional classification setting. This pattern aligns with our ablation findings and suggests that structured pragmatic discourse analysis imposes greater reasoning and instruction-following demands than thematic categorization alone.

\begin{table}[h]
\centering
\footnotesize
\setlength{\tabcolsep}{3.8pt}
\renewcommand{\arraystretch}{1.08}
\caption{Cross-model eval. Each cell reports Acc./$\kappa$.}
\label{tab:cross_model}
\begin{tabular}{llccc}
\toprule
\textbf{Model} & \textbf{Domain} & \textbf{Aspect} & \textbf{Stance} & \textbf{6-Axis} \\
\midrule
\multirow{4}{*}{\shortstack{Gemini\\2.5 Flash}}
  & Ozemp.  & \textbf{79.23 / .77} & \textbf{91.55 / .87} & \textbf{86.15 / .76} \\
  & TRT     & \textbf{89.12 / .88} & \textbf{92.15 / .88} & \textbf{88.62 / .77} \\
  & Collag. & \textbf{86.44 / .85} & \textbf{85.59 / .77} & \textbf{83.97 / .72} \\
  & Fast.   & \textbf{77.64 / .75} & 83.38 / .73          & \textbf{82.23 / .59} \\
\midrule
\multirow{4}{*}{\shortstack{Qwen\\3.5 35B}}
  & Ozemp.  & 77.54 / .75 & 85.26 / .78 & 81.58 / .70 \\
  & TRT     & 79.00 / .77 & 83.30 / .74 & 82.05 / .62 \\
  & Collag. & 81.36 / .79 & 78.39 / .65 & 81.43 / .68 \\
  & Fast.   & 78.25 / .76 & \textbf{86.71 / .79} & 80.72 / .58 \\
\bottomrule
\end{tabular}
\vspace{-10pt}
\end{table}

These results suggest that pragmatic discourse modeling remains substantially more difficult than thematic categorization and stance prediction. Performance degrades under broader contextual settings and high-dimensional pragmatic inference, indicating important limitations in current monolithic prompting approaches. The divergent behavior observed across tasks further suggests that future systems may benefit from task decomposition and specialized inference strategies.

\section{Conclusion}
We introduced the task of structured claim-level discourse analysis for dense health narratives and presented a benchmark for modeling thematic, stance, and pragmatic discourse structure in conversational health content. These domain choices and taxonomies reflect the health-discourse setting studied here and are not intended as a fixed or universal schema. 
Beyond this specific setting, our experiments showed that different discourse tasks require substantially different forms of contextual reasoning. While current LLMs perform strongly on thematic categorization and stance prediction, performance degrades considerably for high-dimensional pragmatic discourse profiling. These findings suggest that future discourse systems may benefit from task decomposition and specialized inference strategies rather than relying on a single monolithic prompting configuration. While developed for health discourse, the underlying representation is not inherently domain-specific and may extend to other dense narrative settings, such as financial advice, political commentary, or legal discourse; we leave full cross-domain validation to future work. We hope this benchmark encourages further research on structured discourse understanding in complex narrative environments.
\section{Limitations}

While our framework and benchmark establish a rigorous foundation for fine-grained claim modeling, several limitations must be acknowledged.

Our study focuses on English-language health discourse collected from YouTube short-form videos within four health domains. Although these domains capture diverse forms of online health communication, the benchmark does not cover the full range of medical topics, platforms, or cultural communication styles present in broader social media ecosystems. In addition, the benchmark uses transcripts as the primary unit of analysis, abstracting away multimodal dimensions of the original videos such as visual demonstrations, on-screen text, and creator affect. In health communication settings, these non-verbal signals can shape claim interpretation, evidential framing, and persuasive intent in ways that transcript-only representations cannot capture. Future work may benefit from integrating visual and paralinguistic signals into the proposed discourse framework.
Our best-performing pragmatic typology results rely on heuristic discourse constraints manually derived from annotation reconciliation sessions(Section~\ref{typology_cls}). These heuristics serve as a proof-of-concept that rule-augmented prompting can substantially improve high-dimensional pragmatic classification, but they were constructed specifically for this benchmark's domains and typology, and their transferability to new domains or label schemas has not been tested. We leave the development of learnable, data-driven constraints that could generalize automatically across domains to future work.

The proposed representation framework focuses on claim-level semantic and pragmatic structure within single-speaker narrative discourse rather than multi-speaker conversational interaction. Consequently, the benchmark does not explicitly model dialogue phenomena such as turn-taking, speaker coordination, or cross-speaker stance dynamics.

\textbf{Ethics Statement}
This benchmark is constructed from publicly available YouTube videos collected via the YouTube Data API v3 for non-commercial research purposes. No private user data was collected. The dataset contains health-related claims sourced from social media that may be medically inaccurate, speculative, or commercially motivated; we do not endorse any claims present in the corpus. Models trained or evaluated on this benchmark should not be deployed in clinical or public health settings without appropriate expert oversight.
GPT-5 was used solely for grammar and proofreading assistance on the manuscript. All research uses of large language models are described in Sections ~\ref{sec:benchmark} and \ref{sec:automation}.

\section{Acknowledgments}
This work was supported in part by the U.S. NSF under awards 2107213, 2026513, and 2309318. We thank Jacob Meltzer for their valuable effort in data preparation and annotation.

\bibliography{custom}

\section{Appendix}

\subsection{Related Work}
\label{sec:relatedwork-appendix}

\textbf{Human--LLM Hybrid Annotation}: 
Recent work increasingly treats large language models as collaborative annotation agents operating within human-supervised workflows rather than fully autonomous labeling systems \cite{li2023coannotating,wang2024human,kim2024meganno+,pangakis2025keeping,jung2024trust,nasution2024chatgpt,zhang-etal-2019-invest}. These studies show that LLMs can substantially accelerate annotation and benchmark construction while still requiring calibration, verification, and expert reconciliation to maintain reliability \cite{dragut-etal-2021-data,findeis2025can,detommaso2024multicalibration,liu2024calibrating,tavakoli2025reliable}.
Our work follows this broader hybrid annotation paradigm but targets substantially richer narrative environments by jointly modeling atomic claims, thematic aspects, stance, and multidimensional pragmatic structure.

\textbf{Health Discourse on Social Media}:
Social media platforms increasingly shape public health communication through decentralized peer, influencer, and consumer-generated narratives rather than traditional clinical authorities \cite{javaid2024trends, fong2024ozempic}. This shift is especially pronounced on video-centric platforms such as YouTube and TikTok, where users share personal experiences, medication advice, side-effects, and persuasive health narratives at scale \cite{yeung2025online, basch2023descriptive}.—environments that studies show amplify anecdotal evidence, off-label medication discourse, and incomplete or misleading health interpretations \cite{javaid2024trends, han2024public, quddos2023semaglutide}.
Existing computational approaches to online health discourse rely largely on coarse topic modeling or sentiment analysis \cite{javaid2024trends, fong2024ozempic}. However, dense health narratives often interleave causal claims, uncertainty, experiences, and persuasive rhetoric within short conversational spans, making flat, document-level representations insufficient for fine-grained analysis. These limitations motivate structured representations capable of modeling claim-level semantic and pragmatic structure \cite{duan2025crowdsourcing}.

\begin{table*}[ht]
\centering
\small
\setlength{\tabcolsep}{8pt}
\renewcommand{\arraystretch}{1.6}
\caption{The Multi-Dimensional Health Claim Taxonomy (6-Axis Typology).}
\label{tab:6_axis_taxonomy}
\begin{tabularx}{\textwidth}{>{\raggedright\arraybackslash}p{4.5cm} X}
\hline
\textbf{Axis \& Definition} & \textbf{Labels} \\
\hline

\rowcolor{gray!10}
\textbf{I. Logical Relationship} \newline {\scriptsize The structural mechanism connecting concepts within the assertion.} & 
{\footnotesize
(1) Causal \textbullet\ (2) Correlational \textbullet\ (3) Comparative \textbullet\ (4) Conditional \textbullet\ (5) None\_Logical 
} \\
\hline

\rowcolor{gray!5}
\textbf{II. Verifiability \& Intent} \newline {\scriptsize The nature of the statement regarding biological reality or logistical context.} & 
{\footnotesize
(1) Factual(Clinical/Biological) \textbullet\ (2) Factual(Logistical/Contextual) \textbullet\ (3) Value(Subjective) \textbullet\ (4) Policy(Prescriptive) \textbullet\ (5) Speculative/Rumor
} \\
\hline

\rowcolor{gray!10}
\textbf{III. Evidence Basis} \newline {\scriptsize The implied source of authority or support used to back the claim.} & 
{\footnotesize
(1) Scientific(Statistical) \textbullet\ (2) Scientific(Expert/Authority) \textbullet\ (3) Scientific(General/Vague) \textbullet\ (4) External Reference(Media/Cultural) \textbullet\ (5) Anecdotal/Personal \textbullet\ (6) Visual/Observable \textbullet\ (7) General Knowledge/Common Sense \textbullet\ (8) Commercial
} \\
\hline

\rowcolor{gray!5}
\textbf{IV. Certainty} \newline {\scriptsize The speaker's confidence level and the use of linguistic hedges.} & 
{\footnotesize
(1) Absolute/Imperative \textbullet\ (2) Hedged(Probabilistic) \textbullet\ (3) Attributed/Distanced
} \\
\hline

\rowcolor{gray!10}
\textbf{V. Temporality} \newline {\scriptsize The time orientation or universal principle of the assertion.} & 
{\footnotesize
(1) Current/Past \textbullet\ (2) Counterfactual \textbullet\ (3) Future/Predictive \textbullet\ (4) Sequential \textbullet\ (5) Timeless
} \\
\hline

\rowcolor{gray!5}
\textbf{VI. Claim Focus} \newline {\scriptsize The primary rhetorical intent and structural purpose of the claim.} & 
{\footnotesize
(1) Explanatory/Argumentative \textbullet\ (2) Descriptive/Definitional \textbullet\ (3) Narrative/Sequential \textbullet\ (4) Evaluative \textbullet\ (5) Promotional / Commercial
} \\
\hline
\end{tabularx}
\end{table*}

\subsection{Multi-Dimensional Health Claim Taxonomy}
\label{app:6_axis_taxonomy_sec}

This section details the six-dimensional pragmatic typology used to analyze rhetorical and communicative features in dense health narratives. Table~\ref{tab:6_axis_taxonomy} provides the formalized definitions, categorical labels, and structural scopes for each of the orthogonal axes.

\subsection{Extended Qualitative Analysis of Multi-Aspect Entanglement and Rhetorical Trajectories}
\label{app:pragmatic_divergence}

To complement the paired-claim comparison presented in the main text, this subsection provides broader qualitative evidence demonstrating why both domain-specific aspect taxonomies and multi-axis pragmatic profiling are required to capture the complexity of health discourse.

\noindent\textbf{Multi-Aspect Entanglement in Dense Narratives.} 
As discussed in Section~\ref{Introduction}, health discourse exhibits high claim density (averaging 13.22 claims per minute), wherein speakers rapidly interleave assertions across multiple distinct health dimensions with opposing stances. Coarser representations, such as document-level sentiment or broad topic categorization, completely collapse these distinctions. Table~\ref{tab:multi_aspect_entanglement} illustrates this phenomenon using a representative transcript span from the TRT domain, showing how a single speaker simultaneously moves across muscle performance, adverse side effects, sexual health, severe medical risks, and clinical eligibility within seconds.

\begin{table*}[t]
\centering
\caption{Representative transcript excerpt demonstrating multi-aspect entanglement, where a single conversational span interleaves multiple health dimensions and opposing stances.}
\small
\setlength{\tabcolsep}{5pt}
\begin{tabular}{p{9.2cm}|p{4.5cm}|c}
\hline
\textbf{Claim (Transcript Excerpt)} & \textbf{Thematic Aspect} & \textbf{Stance} \\
\hline
``you'll look better, you'll feel better, and you'll put muscle on faster'' & Muscle, Performance \& Aesthetics & Positive \\
\hline
``will f**k your hormonal system up indefinitely if you do it too long'' & Side Effects \& Adverse Reactions & Negative \\
\hline
``they have a lot of problems with getting erections'' & Sexual Health \& Libido & Negative \\
\hline
``they have liver problems'' & Medical Health \& Biomarkers & Negative \\
\hline
``Many of them die from heart attacks early'' & Severe Risks \& Long-Term Safety & Negative \\
\hline
``I still got washboard abs, I can still do single-arm chin-ups'' & Muscle, Performance \& Aesthetics & Positive \\
\hline
``the only place for it is when you've got four doctors in balance, you've detoxified your body'' & Clinical Diagnosis \& Patient Eligibility & Neutral \\
\hline
\end{tabular}

\label{tab:multi_aspect_entanglement}
\end{table*}

\noindent\textbf{Intra-Speaker Rhetorical Strategy Trajectories.} 
Beyond aspect-level assignment, capturing pragmatic variation across multiple axes reveals structural shifts in a speaker's rhetorical strategy that remain invisible under flat sentiment or stance frameworks. To demonstrate this, Table\mbox{~\ref{tab:rhetorical_trajectory}} tracks six consecutive claims made by a single physician discussing cardiovascular risk under predominantly the same thematic aspect (\textit{Severe Risks \& Long-Term Safety}).

The pragmatic profile shifts dynamically across the narrative: the speaker opens with a categorical framing statement, transitions to public concern, cites published expert literature, summarizes a trend across studies, describes their own research methodology, and finally concludes with a general statement. This complete rhetorical arc, from initial framing through evidence accumulation to final assertion, proves the necessity of decomposing discourse into multi-dimensional pragmatic axes.

\begin{table*}[t]
\centering
\caption{Intra-speaker rhetorical strategy trajectory across six claims by a single physician predominantly under the same thematic aspect (\textit{Severe Risks}), illustrating how the 6-axis typology captures shifts in evidential framing, certainty, and logical structure that stance alone cannot represent.}
\tiny
\setlength{\tabcolsep}{2.5pt}
\resizebox{\textwidth}{!}{
\begin{tabular}{p{3.2cm} p{1.5cm} c p{1.1cm} p{1.4cm} p{1.6cm} p{1.1cm} p{1.2cm} p{1.1cm}}
\toprule
\textbf{Claim} & \textbf{Aspect} & \textbf{Stance} & \textbf{Logical} & \textbf{Verif.} & \textbf{Evidence} & \textbf{Certainty} & \textbf{Temp.} & \textbf{Focus} \\
\midrule
``Cardiovascular risk is a subject of controversy regarding TRT.'' & Severe Risks & Neg & Non\_Log & Factual (Clin.) & Sci. (Gen./Vague) & Absolute & Current/Past & Descriptive \\
\addlinespace[0.3em]
``There was significant concern that testosterone may cause a heart attack.'' & Severe Risks & Neg & Causal & Factual (Clin.) & General Knowledge & Hedged & Current/Past & Descriptive \\
\addlinespace[0.3em]
``Men with lower testosterone were significantly more likely to die earlier'' (citing Molly Shores, 2006). & Severe Risks & Neu & Comparative & Factual (Clin.) & Sci. (Expert/Auth.) & Attributed & Current/Past & Descriptive \\
\addlinespace[0.3em]
``Men with lower testosterone exhibited a general trend of higher death rates.'' & Severe Risks & Pos & Correlational & Factual (Clin.) & Sci. (Gen./Vague) & Hedged & Current/Past & Descriptive \\
\addlinespace[0.3em]
``We found over 200 articles addressing testosterone and cardiovascular disease.'' & Medical Health & Neu & Non\_Log & Factual (Log.) & Anecdotal/Personal & Attributed & Current/Past & Narrative \\
\addlinespace[0.3em]
``Low testosterone is a risk factor for cardiovascular events.'' & Severe Risks & Neg & Correlational & Factual (Clin.) & Sci. (Gen./Vague) & Absolute & Timeless & Descriptive \\
\bottomrule
\end{tabular}
}
\label{tab:rhetorical_trajectory}
\end{table*}

\subsection{Transcript Triage and Filtering Prompt}
\label{app:triage_prompt}

This prompt template was used for initial screening of video transcripts harvested from the YouTube Data API. This triage phase ensures the relevance and linguistic consistency of the candidate pool before purposive sampling. Placeholders \texttt{\{TOPIC\}} and \texttt{\{transcript\_text\}} are populated at runtime. 

\begin{tcolorbox}[
    breakable,
    enhanced,
    colback=gray!5,
    colframe=gray!50,
    boxrule=0.4pt,
    arc=1pt,
    left=4pt,
    right=4pt,
    top=4pt,
    bottom=4pt,
    boxsep=1pt,
    fontupper=\fontsize{7.5}{8.5}\selectfont\ttfamily,
    title={\scriptsize\sffamily Transcript Triage and Filtering Prompt Template},
    label=box:triage_prompt
]
\begin{flushleft}
You are an AI assistant performing an initial triage on the transcript of a YouTube video. Your task is to quickly determine if the transcript is usable and relevant to the topic of interest.

Topic of Interest: \{TOPIC\}

Transcript:
\end{flushleft}
\begin{verbatim}
{transcript_text}
\end{verbatim}
\begin{flushleft}
\#\#\# INSTRUCTIONS:

Respond ONLY with a valid JSON object containing three keys:
1. has\_meaningful\_speech (boolean): true if the transcript has coherent sentences.
2. is\_english (boolean): true if the primary language is English.
3. is\_relevant (boolean): true if the transcript discusses the Topic of Interest, with a brief relevance\_reason (string).

Response Format
\end{flushleft}
\begin{verbatim}
{
  "has_meaningful_speech": true,
  "is_english": true,
  "is_relevant": true,
  "relevance_reason": 
    "Explain briefly why the content is on-topic"
}
\end{verbatim}
\end{tcolorbox}

\label{app:denoising_prompt}

\subsection{Orthographic Denoising Evaluation}
\label{app:denoising_examples}

We evaluated all 60 videos using two complementary metrics adapted from prior ASR evaluationquality assessment work: Medical Term Accuracy 
(MTA)~\cite{shor2023clinical} and Punctuation Quality (PQ)~\cite{meister2023librispeech}. Both metrics were computed on the raw Whisper transcripts and the transcripts produced after the output and \emph{Orthographic Denoising and Stabilization} stage to enable direct comparison. As shown in Table~\ref{tab:transcript_quality}, denoising improved average Medical Term Accuracy (MTA) from 0.97 to 1.0 and average PQ from 0.95 to 0.98 across the full corpus. The normalization stage proved critical for a non-trivial subset of transcripts.
Five videos exhibited raw MTA at or below 0.67, all five recovered to 1.0 following denoising. Similarly, five videos exhibited raw PQ below 0.9, with three falling below 0.40, recovering to a mean of 0.94 post-denoising, with the most severely degraded transcript improving from 0.22 to 0.97. Tables~\ref{tab:denoise_ex1} and~\ref{tab:denoise_ex2} present representative corrections.

\begin{table}[h]
\centering
\caption{Medical Term Accuracy and Punctuation Quality scores for raw Whisper ASR output and post-denoising transcripts ($N=60$).}
\small
\setlength{\tabcolsep}{2pt}
\begin{tabular}{lcccccc}
\toprule
\textbf{Domain} 
& \textbf{\shortstack{Med.\\Acc.\\(Raw)}} 
& \textbf{\shortstack{Med.\\Acc.\\(Den.)}} 
& \textbf{\shortstack{$\Delta$\\Med.}} 
& \textbf{\shortstack{Punct.\\(Raw)}} 
& \textbf{\shortstack{Punct.\\(Den.)}} 
& \textbf{\shortstack{$\Delta$\\Punct.}} \\
\midrule
Ozempic  & 0.889 & 1.000 & $+$0.111 & 0.894 & 0.983 & $+$0.089 \\
TRT      & 1.000 & 1.000 & $+$0.000 & 0.958 & 0.975 & $+$0.017 \\
Collagen & 1.000 & 1.000 & $+$0.000 & 0.969 & 0.967 & $-$0.002 \\
Fasting  & 0.965 & 1.000 & $+$0.035 & 0.971 & 0.973 & $+$0.002 \\
\midrule
\textbf{Overall} & \textbf{0.963} & \textbf{1.000} & $\mathbf{+}$\textbf{0.037} & \textbf{0.948} & \textbf{0.975} & $\mathbf{+}$\textbf{0.027} \\
\bottomrule
\end{tabular}

\label{tab:transcript_quality}
\end{table}

\begin{table}[h]
\centering
\caption{Denoising example from the Ozempic domain}
\small
\setlength{\tabcolsep}{4pt}
\renewcommand{\arraystretch}{1.1}
\begin{tabular}{p{0.46\linewidth} p{0.46\linewidth}}
\toprule
\textbf{Raw Whisper Output} & \textbf{Denoised Output} \\
\midrule

\ldots heard about the buzz around
\colorbox{yellow}{a Zempic} Weight Loss.
&
\ldots heard about the buzz around
\colorbox{yellow}{Ozempic} weight loss. \\

\ldots the weight \colorbox{yellow}{all he} seems to come back.
&
\ldots the weight \colorbox{yellow}{always} seems to come back. \\

\ldots that \colorbox{yellow}{a Zempic} might seem like a quick fix \ldots
&
\ldots that \colorbox{yellow}{Ozempic} might seem like a quick fix \ldots \\

\ldots beyond just \colorbox{yellow}{temperate to temperate} discomfort.
&
\ldots beyond just \colorbox{yellow}{temporary to temporary} discomfort. \\

\bottomrule
\end{tabular}

\label{tab:denoise_ex1}
\end{table}

\begin{table}[h]
\centering
\caption{Denoising example from the Fasting domain}
\small
\setlength{\tabcolsep}{4pt}
\renewcommand{\arraystretch}{1.1}
\begin{tabular}{p{0.46\linewidth} p{0.46\linewidth}}
\toprule
\textbf{Raw Whisper Output} & \textbf{Denoised Output} \\
\midrule

\ldots something called \colorbox{yellow}{a topogy}, which is \ldots
&
\ldots something called \colorbox{yellow}{autophagy}, which is \ldots \\

\colorbox{yellow}{Now what} the studies show is that fasting triggered
\colorbox{yellow}{a topogy} in the brain
&
\colorbox{yellow}{Now,} what the studies show is that fasting-triggered
\colorbox{yellow}{autophagy} \ldots \\

\colorbox{yellow}{So really} great for brain...
&
\colorbox{yellow}{So,} really great for brain... \\

\ldots to reduce \colorbox{yellow}{the body-wide} inflammation.
&
\ldots to reduce \colorbox{yellow}{body-wide} inflammation. \\

\colorbox{yellow}{So try} maybe fasting for 24 hours.
&
\colorbox{yellow}{So,} try maybe fasting for 24 hours. \\

\bottomrule
\end{tabular}

\label{tab:denoise_ex2}
\end{table}

\subsection{Orthographic Denoising and Stabilization}
\label{app:denoising_prompt_template}
The following prompt template was used with Gemini 2.5 Flash to stabilize raw Whisper ASR output. 

\begin{tcolorbox}[
    breakable,
    enhanced,
    colback=gray!5,
    colframe=gray!50,
    boxrule=0.4pt,
    arc=1pt,
    left=4pt,
    right=4pt,
    top=4pt,
    bottom=4pt,
    boxsep=1pt,
    fontupper=\fontsize{7.5}{8.5}\selectfont\ttfamily,
    title={\scriptsize\sffamily Orthographic Denoising Prompt Template},
    label=box:denoising_prompt
    ]
\begin{flushleft}
You are an expert-level text editor and proofreader specializing in medical transcripts. The following text is an AI-generated transcript from Whisper. It likely contains spelling errors and incorrect punctuation.

Topic of Interest: \{TOPIC\}

Transcript:
\end{flushleft}
\begin{verbatim}
{transcript_text}
\end{verbatim}
\begin{flushleft}
\#\#\# INSTRUCTIONS:

1. Correct Spelling: Fix spelling errors. Pay special attention to metabolic, physiological, and topic-specific terms. 
(Example keywords provided for \{TOPIC\} included: [e.g., Autophagy, Insulin Sensitivity, Glycogen, Bioavailability, etc.])

2. Fix Punctuation: Fix capitalization, sentence breaks, and punctuation to create clean, readable text.

3. Conservative Editing: Do NOT summarize. Do NOT delete content. Do NOT add new content. Keep the text as close to the original meaning as possible, just cleaned up.

Response Format

Respond ONLY with a valid JSON object containing a single key:
\end{flushleft}
\begin{verbatim}
{
  "denoised_transcript": 
    "The corrected and punctuated text here..."
}
\end{verbatim}
\end{tcolorbox}

\subsection{Annotation Labor and Complexity}
\label{app:annot}
Table~\ref{tab:annotation_effort} summarizes the approximate expert effort required to construct the gold-standard benchmark. The reported estimates include both independent annotation and collaborative reconciliation phases across the full multidimensional labeling workflow.

\begin{table}[h]
\centering
\small
\setlength{\tabcolsep}{4pt} 
\renewcommand{\arraystretch}{1.2}
\caption{Expert labor intensity per video (Cumulative Person-Minutes).}
\label{tab:annotation_effort}
\begin{tabularx}{\columnwidth}{X r} 
\hline
\textbf{Annotation Phase} & \textbf{Avg. Person-Mins} \\ \hline
\textit{Independent Phase (For 2 Annotators)} & \\ 
\hspace{3mm}Statement Extraction & 10--20 min \\
\hspace{3mm}Atomic Decomposition & 30--50 min \\
\hspace{3mm}Multidimensional Labeling & 60--90 min \\ \hline
\textit{Consensus Phase} & \\ 
\hspace{3mm}Consensus \& Dispute Resolution & 25--40 min \\ \hline
\textbf{Total Expert Time per Video} & \textbf{125--200 min} \\ \hline
\end{tabularx}
\end{table}

\begin{table*}[t]
\centering
\small
\setlength{\tabcolsep}{8pt} 
\renewcommand{\arraystretch}{1.2}
\caption{Inter-annotator agreement metrics across hierarchical pipeline layers.}
\label{tab:iaa_master}
\begin{tabularx}{\textwidth}{Xcccccc}
\hline
\textbf{Annotation Layer / Axis} & \textbf{Metric} & \textbf{Ozempic} & \textbf{TRT} & \textbf{Collagen} & \textbf{Fasting} & \textbf{Mean} \\ \hline
Statement Extraction      & Pairwise $F_1$  & 0.78 & 0.99 & 0.95 & 0.98 & 0.93 \\
Atomic Decomposition       & Raw Agreement   & 0.89 & 0.93 & 0.92 & 0.95 & 0.92 \\ \hline
\textbf{Aspect Assignment} & Cohen's $\kappa$ & 0.89 & 0.81 & 0.88 & 0.75 & 0.83 \\
                          & Raw Agreement   & 0.90 & 0.83 & 0.89 & 0.78 & 0.85 \\ \hline
\textbf{Stance Classification} & Cohen's $\kappa$ & 0.65 & 0.87 & 0.84 & 0.83 & 0.80 \\
                          & Raw Agreement   & 0.77 & 0.92 & 0.90 & 0.89 & 0.87 \\ \hline
\noalign{\smallskip}
\multicolumn{7}{l}{\textbf{6-Axis Typology Dimensions}} \\
\noalign{\smallskip}
~~Axis I: Logical Relationship     & Cohen's $\kappa$ & 0.83 & 0.86 & 0.88 & 0.82 & 0.85 \\
                                  & Raw Agreement   & 0.90 & 0.91 & 0.92 & 0.89 & 0.90 \\
~~Axis II: Verifiability \& Intent & Cohen's $\kappa$ & 0.63 & 0.74 & 0.82 & 0.81 & 0.75 \\
                                  & Raw Agreement   & 0.79 & 0.85 & 0.90 & 0.89 & 0.86 \\
~~Axis III: Evidence Basis         & Cohen's $\kappa$ & 0.64 & 0.71 & 0.90 & 0.85 & 0.77 \\
                                  & Raw Agreement   & 0.76 & 0.87 & 0.92 & 0.95 & 0.88 \\
~~Axis IV: Certainty                & Cohen's $\kappa$ & 0.79 & 0.68 & 0.80 & 0.88 & 0.79 \\
                                  & Raw Agreement   & 0.92 & 0.89 & 0.94 & 0.97 & 0.93 \\
~~Axis V: Temporality              & Cohen's $\kappa$ & 0.77 & 0.65 & 0.88 & 0.67 & 0.74 \\
                                  & Raw Agreement   & 0.87 & 0.86 & 0.95 & 0.90 & 0.89 \\
~~Axis VI: Claim Focus            & Cohen's $\kappa$ & 0.82 & 0.56 & 0.81 & 0.53 & 0.68 \\
                                  & Raw Agreement   & 0.86 & 0.75 & 0.87 & 0.71 & 0.80 \\ \hline
\textit{6-Axis Typology (Mean)}   & Cohen's $\kappa$ & 0.75 & 0.70 & 0.85 & 0.76 & 0.76 \\
                                  & Raw Agreement   & 0.85 & 0.85 & 0.92 & 0.88 & 0.88 \\ \hline
\end{tabularx}
\end{table*}

\begin{table*}[ht]
\centering
\small
\setlength{\tabcolsep}{4pt} 
\renewcommand{\arraystretch}{1.45}
\caption{Canonical aspect taxonomies induced across all four health domains.}
\label{tab:aspect_taxonomies_table}
\begin{tabularx}{\textwidth}{>{\bfseries}l X}
\hline
\textbf{Topic} & \textbf{Aspects} \\
\hline

\rowcolor{blue!6}
Ozempic &
{\footnotesize
(1) Weight Loss Effectiveness \textbullet\ 
(2) Medical Health Benefits (Non-Weight) \textbullet\ 
(3) Appetite \& Satiety \textbullet\ 
(4) Gastrointestinal \& Acute Side Effects \textbullet\ 
(5) Mental \& Emotional Impact \textbullet\ 
(6) Aesthetic \& Physical Transformation \textbullet\ 
(7) Financial \& Insurance \textbullet\ 
(8) Logistics \& Supply Chain \textbullet\ 
(9) Medical Regimen \& Dosing \textbullet\ 
(10) Dietary \& Lifestyle Changes \textbullet\ 
(11) Long-term Safety \& Risks \textbullet\ 
(12) Social Stigma \& Perception \textbullet\ 
(13) Product Comparisons \& Alternatives
} \\
\hline

\rowcolor{green!6}
TRT &
{\footnotesize
(1) Muscle, Performance \& Aesthetics \textbullet\ 
(2) Mental \& Cognitive Impact \textbullet\ 
(3) Sexual Health \& Libido \textbullet\ 
(4) Energy \& Vitality \textbullet\ 
(5) Medical Health \& Biomarkers \textbullet\ 
(6) Side Effects \& Adverse Reactions \textbullet\ 
(7) Fertility \& HPTA Shutdown \textbullet\ 
(8) Dosing Schedule \& Protocol Strategy \textbullet\ 
(9) Medication Form \& Delivery Tools \textbullet\ 
(10) Social Stigma \& Perception \textbullet\ 
(11) Financial, Market \& Access \textbullet\ 
(12) Severe Risks \& Long-Term Safety \textbullet\ 
(13) Lifestyle \& Holistic Integration \textbullet\ 
(14) Clinical Diagnosis \& Patient Eligibility
} \\
\hline

\rowcolor{orange!8}
Collagen & 
{\footnotesize
(1) Ingredients \& Formulation \textbullet\ 
(2) Sensory \& Usability \textbullet\ 
(3) Quality \& Safety \textbullet\ 
(4) Skin Health \& Anti-Aging \textbullet\ 
(5) Joint \& Bone Mobility \textbullet\ 
(6) Hair \& Nail Fortification \textbullet\ 
(7) Muscle Growth \& Body Composition \textbullet\ 
(8) Digestion \& Satiety \textbullet\ 
(9) General Wellbeing \textbullet\ 
(10) Scientific Validation \textbullet\ 
(11) Price \& Value \textbullet\ 
(12) Brand Reputation \& Market
} \\
\hline

\rowcolor{purple!8}
Fasting & 
{\footnotesize
(1) Weight Loss \& Body Composition \textbullet\ 
(2) Metabolism \& Energy Processing \textbullet\ 
(3) Cellular Health \& Longevity \textbullet\ 
(4) Hormonal Balance \& Endocrine Health \textbullet\ 
(5) Appetite, Hunger \& Cravings \textbullet\ 
(6) Mental Health \& Cognitive Function \textbullet\ 
(7) Physical Energy \& Vitality \textbullet\ 
(8) Immunity \& Inflammation \textbullet\ 
(9) Organ \& Cardiovascular Health \textbullet\ 
(10) Fasting Protocols \& Timing Strategy \textbullet\ 
(11) Nutritional Intake \& Food Choices \textbullet\ 
(12) Lifestyle Convenience \& Flexibility \textbullet\ 
(13) Medical Guidance, Safety \& Scientific Evidence \textbullet\ 
(14) Evolutionary \& Psychological Responses \textbullet\ 
(15) Side Effects \& Adverse Reactions \textbullet\ 
(16) Sleep \& Recovery \textbullet\ 
(17) Social Perception \& Public Awareness \textbullet\ 
(18) Market, Industry \& Financial Access \textbullet\ 
(19) Adjunct Fitness \& Holistic Habits
} \\
\hline

\end{tabularx}
\end{table*}

\subsection{Aspect Taxonomy Instantiation}
\label{app:aspect_taxonomies}
Table~\ref{tab:aspect_taxonomies_table} presents the canonical aspect taxonomies induced across all four health domains.

\subsection{Structured Discourse Prediction Prompts}
\label{app:labeling_prompt}

We used two prompt templates for automated structured discourse prediction: one for joint aspect and stance classification, and one for six-dimensional pragmatic typology classification. Both prompts share a fixed instruction structure across all four domains; domain-specific content (aspect taxonomy, definitions, keywords, and few-shot examples) is populated via placeholders at runtime. 
The aspect taxonomy used to populate \texttt{\{ASPECT\_TAXONOMY\}} is provided in full in Table~\ref{tab:aspect_taxonomies_table}.

\paragraph{Prompt 1: Aspect and Stance Classification.}
\leavevmode\vspace{0.5em}
\begin{tcolorbox}[
    breakable,
    enhanced,
    colback=gray!5,
    colframe=gray!50,
    boxrule=0.4pt,
    arc=1pt,
    left=4pt,
    right=4pt,
    top=4pt,
    bottom=4pt,
    boxsep=1pt,
    fontupper=\fontsize{7.5}{8.5}\selectfont\ttfamily,
    title={\scriptsize\sffamily Aspect \& Stance Classification Prompt Template},
    label=box:aspect_stance_prompt
]
\begin{flushleft}
You are a senior health communication researcher. Analyze the following list of ``Target Claims'' about the Topic of Interest.

Topic of Interest: \{TOPIC\}

\#\#\# TAXONOMY: \{N\} MASTER ASPECTS

\{ASPECT\_TAXONOMY\}

[Domain-specific aspect labels, definitions, and keywords. See Table~\ref{tab:aspect_taxonomies_table} for all four domains.]

\#\#\# STANCE DEFINITIONS
\begin{itemize}[leftmargin=*, nosep]
  \item Positive: The claim describes a benefit, a success, a favorable outcome, or an improvement.
  \item Negative: The claim describes a harm, a failure, a side effect, a risk, or an unfavorable outcome.
  \item Neutral: The claim is a factual observation, a description of medical protocol, or a statement of intent/action without a positive or negative judgment.
\end{itemize}

\#\#\# TARGET CLAIMS LIST:
\end{flushleft}
\begin{verbatim}
{CLAIMS_LIST_JSON}
\end{verbatim}
\begin{flushleft}
\#\#\# INSTRUCTIONS:
1. For each claim, first provide a 1-sentence Reasoning identifying the core subject and context.
2. Based on that reasoning, assign exactly ONE Aspect Label and ONE Stance.
3. Respond ONLY with a valid JSON object containing a list named ``classified\_claims''.

Response Format
\end{flushleft}
\begin{verbatim}
{
  "classified_claims": [
    {
      "claim_id": 0,
      "reasoning": "Explain the medical or
      logistical context",
      "aspect_label": "Label Name",
      "stance": "Sentiment"
    }
  ]
}
\end{verbatim}
\end{tcolorbox}

\paragraph{Prompt 2: Six-Dimensional Pragmatic Typology Classification.}
\leavevmode\vspace{0.5em}
\begin{tcolorbox}[
    breakable,
    enhanced,
    colback=gray!5,
    colframe=gray!50,
    boxrule=0.4pt,
    arc=1pt,
    left=4pt,
    right=4pt,
    top=4pt,
    bottom=4pt,
    boxsep=1pt,
    fontupper=\fontsize{7.5}{8.5}\selectfont\ttfamily,
    title={\scriptsize\sffamily 6-Axis Typology Classification Prompt Template},
    label=box:typology_prompt
]
\begin{flushleft}
You are a senior health communication researcher. Analyze the ``Target Claim'' provided about \{TOPIC\}.

To help you understand the intent, I have provided the Surrounding Context (the 2 claims immediately before and after this claim in the video).

IMPORTANT GENERAL RULES
\begin{itemize}[leftmargin=*, nosep]
  \item Always pick the single best-fitting label for each axis (no multi-labeling).
  \item Use the Surrounding Context to clarify ambiguity, but label ONLY the ``Target Claim.''
  \item If a statement contains multiple parts, label the main claim (the core assertion).
  \item Be consistent across items: similar phrasing should yield similar labels.
\end{itemize}

[Axis definitions for I--VI as specified in Table~\ref{tab:6_axis_taxonomy}]

\#\#\# GOLD STANDARD FEW-SHOT EXAMPLES

\{FEW\_SHOT\_EXAMPLES\}

[5 manually annotated claims drawn from the target domain. Examples vary per domain to reflect domain-specific vocabulary]

\#\#\# SURROUNDING CONTEXT:

\{SURROUNDING\_CONTEXT\}

{>}{>}{>} TARGET CLAIM TO ANALYZE: ``\{TARGET\_CLAIM\}'' \textless

Response Format
\end{flushleft}
\begin{verbatim}
{
  "Logical": "Label",
  "Verifiability": "Label",
  "Evidence Basis": "Label",
  "Certainty": "Label",
  "Temporality": "Label",
  "Claim Focus": "Label"
}
\end{verbatim}
\end{tcolorbox}
\noindent The axis definitions and label inventories supplied to 
Prompt~2 are identical across all four domains and correspond 
directly to the six-dimensional typology defined in 
Table~\ref{tab:6_axis_taxonomy}.

\subsection{Qualitative Error Analysis of Chain-of-Thought Prompting}
\label{sec:cot_error_appendix}

To better understand why a straightforward, single-pass Chain-of-Thought (CoT) prompting underperforms on high-dimensional pragmatic profiling compared to heuristic constraints (Section~\ref{typology_cls}), we performed a qualitative error analysis comparing predictions where heuristic prompting successfully matched the ground truth but CoT failed. This analysis reveals three distinct behavioral failure modes in standard CoT prompting:

\noindent\textbf{1. Holistic Interference from Joint Prediction:} 
Our CoT implementation utilized a lightweight reasoning constraint designed to capture speaker intent across all six pragmatic axes simultaneously. This holistic approach caused structural interference: the generated reasoning would often optimize for one axis (e.g., Claim Focus) while completely misaligning with another (e.g., Logical Relationship).

\noindent\textbf{2. Semantic Over-Analysis over Structural Mapping:} 
The CoT prompt frequently caused the model to over-analyze conversational nuances rather than enforce schema constraints. As shown in Table~\ref{tab:cot_error_examples}, for the claim \textit{``ozempic is the ideal weight loss solution for diabetics,''} the model generates an accurate semantic explanation of the speaker's intent in its reasoning block (positioning it as the ``best possible option''), but completely fails to map that understanding to the correct categorical schema label, defaulting to \textit{None\_logical} instead of \textit{Comparative}.

\noindent\textbf{3. Linguistic Triggers and Rationalization Traps:} 
CoT proved highly vulnerable to modal verbs and ambiguous phrasing. For claims like \textit{``we can use ozempic for weight loss,''} the model got tripped up by surface-level linguistic triggers like the modal verb ``can,'' causing it to ignore the pragmatic context and absolute reality of the health narrative, incorrectly driving an absolute assertion into a \textit{Hedged} classification. Furthermore, on borderline claims (e.g., \textit{``GLP active definitely worth the money''}), the model demonstrated ``rationalization flip-flopping'': because text generation is autoregressive, once the model randomly generated its first few reasoning tokens in a certain direction, it forced itself to write a highly convincing post-hoc rationalization for an incorrect label (fluctuating between \textit{Anecdotal} and \textit{General Knowledge}).

\begin{table*}[h]
\centering
\caption{Representative qualitative error cases illustrating behavioral failure modes of Chain-of-Thought (CoT) prompting on high-dimensional pragmatic discourse classification using Gemini 2.5 Flash.}
\small
\setlength{\tabcolsep}{4pt}
\begin{tabular}{p{3.2cm}|p{2.2cm}|p{2.5cm}|p{2.5cm}|p{4.8cm}}
\hline
\textbf{Target Claim} & \textbf{Dimension} & \textbf{Gold Label} & \textbf{CoT Prediction} & \textbf{Observed Failure Mechanism} \\
\hline
\textit{``Ozempic is the ideal weight loss solution for diabetics.''} & Logical Relationship & Comparative & None\_Logical & \textbf{Semantic Over-Analysis:} Accurately explains the superlative intent in reasoning, but fails to map to the categorical schema label. \\
\hline
\textit{``We can use ozempic for weight loss.''} & Certainty & Absolute / Imperative & Hedged (Probabilistic) & \textbf{Surface Linguistic Triggers:} Tripped up by the modal verb ``can'', misinterpreting an absolute statement as a probabilistic hedge. \\
\hline
\textit{``GLP active definitely worth the money.''} & Evidence Basis & General Knowledge / Common Sense & Anecdotal / Personal & \textbf{Rationalization Flip-Flopping:} Autoregressive generation traps the model into writing post-hoc justifications that fluctuate across runs. \\
\hline
\end{tabular}
\label{tab:cot_error_examples}
\end{table*}

\begin{table*}[!htbp]
\centering
\small
\setlength{\tabcolsep}{10pt}
\caption{Comparison between original submitted evaluation results and the new clean execution baseline on the TRT domain dataset using Gemini 2.5 Flash.}
\begin{tabular}{l|c|c|c}
\hline
\textbf{Evaluation Task} & \textbf{Original Submitted (TRT)} & \textbf{New Baseline ($\text{Mean} \pm \text{SD}$)} & \textbf{Variance} \\
\hline
Thematic Aspect Labeling & 89.12\% & 92.82\% $\pm$ 1.23\% & $\Delta$ +3.70\% \\
Core Stance Detection & 92.15\% & 93.92\% $\pm$ 1.16\% & $\Delta$ +1.77\% \\
6-Axis Typology Mean & 88.62\% & 93.41\% $\pm$ 0.15\% & $\Delta$ +4.79\% \\
\hline
\end{tabular}
\label{tab:baseline_comparison}
\end{table*}
\begin{table*}[!htbp]
\centering
\small
\setlength{\tabcolsep}{3.0pt}
\renewcommand{\arraystretch}{1.2}
\caption{Phase II Optimization: Impact of Heuristic Refinement across Contextual Windows.}
\label{tab:6axis_phase2_mix}
\begin{tabular}{lcccccccccccc}
\toprule
 & \multicolumn{2}{c}{\textbf{Atomic}} 
 & \multicolumn{2}{c}{\textbf{Batch}} 
 & \multicolumn{2}{c}{\textbf{Local $\pm 2$}} 
 & \multicolumn{2}{c}{\textbf{Narr.}} 
 & \multicolumn{2}{c}{\textbf{Trans.}} 
 & \multicolumn{2}{c}{\textbf{Summ.}} \\
\cmidrule(lr){2-3}\cmidrule(lr){4-5}\cmidrule(lr){6-7}
\cmidrule(lr){8-9}\cmidrule(lr){10-11}\cmidrule(lr){12-13}
\textbf{Axis} 
& Acc. & $\kappa$ 
& Acc. & $\kappa$ 
& Acc. & $\kappa$ 
& Acc. & $\kappa$ 
& Acc. & $\kappa$ 
& Acc. & $\kappa$ \\
\midrule
I. Logical Rel.   
& 84.59 & 0.7383 
& 85.30 & \textbf{0.7524} 
& 84.23 & 0.7366 
& 84.59 & 0.7423 
& 82.44 & 0.7012 
& \textbf{85.30} & 0.7507 \\

II. Verifiability 
& 84.95 & 0.7277 
& 84.59 & 0.7303 
& \textbf{86.38} & \textbf{0.7609} 
& 86.38 & 0.7572 
& 84.59 & 0.7209 
& 86.38 & 0.7593 \\

III. Evidence     
& 82.44 & 0.7063 
& \textbf{85.66} & \textbf{0.7616} 
& 84.23 & 0.7391 
& 70.61 & 0.5680 
& 69.89 & 0.5575 
& 77.06 & 0.6493 \\

IV. Certainty     
& 91.04 & 0.7688 
& 91.76 & 0.7713 
& 90.32 & 0.7533 
& 89.96 & 0.7339 
& 92.83 & 0.8101 
& \textbf{93.55} & \textbf{0.8221} \\

V. Temporality    
& 89.96 & 0.8161 
& 90.32 & 0.8283 
& \textbf{91.40} & \textbf{0.8443} 
& 90.32 & 0.8252 
& 89.96 & 0.8181 
& 90.68 & 0.8301 \\

VI. Claim Focus   
& 59.86 & 0.4918 
& 60.93 & 0.5073 
& \textbf{82.08} & \textbf{0.7442} 
& 62.37 & 0.5184 
& 57.71 & 0.4655 
& 56.27 & 0.4638 \\
\midrule
\textbf{Mean} 
& \textbf{82.14} & \textbf{0.7082} 
& \textbf{83.09} & \textbf{0.7252} 
& \textbf{86.44} & \textbf{0.7631} 
& \textbf{80.70} & \textbf{0.6908} 
& \textbf{79.57} & \textbf{0.6789} 
& \textbf{81.54} & \textbf{0.7125} \\
\bottomrule
\end{tabular}
\vspace{-10pt}
\end{table*}

\begin{table*}[!htbp]
\centering
\caption{Full multi-temperature robustness breakdown across 12 execution runs (3 trials per temperature setting) on the TRT domain dataset using Gemini 2.5 Flash.}
\small
\setlength{\tabcolsep}{5pt}
\begin{tabular}{l|c|c|c|c}
\hline
\textbf{Task / Dimension} & \textbf{Tier A ($\tau = 0.0$)} & \textbf{Tier B ($\tau = 0.2$)} & \textbf{Tier C ($\tau = 0.5$)} & \textbf{Tier D ($\tau = 0.8$)} \\
\hline
Aspect & 92.82\% $\pm$ 1.23\% & 92.92\% $\pm$ 0.26\% & 91.68\% $\pm$ 1.52\% & 91.49\% $\pm$ 0.53\% \\
Stance & 93.92\% $\pm$ 1.16\% & 94.18\% $\pm$ 0.11\% & 93.06\% $\pm$ 1.49\% & 93.62\% $\pm$ 0.45\% \\
Logical Relationship & 95.23\% $\pm$ 0.06\% & 95.18\% $\pm$ 0.14\% & 95.35\% $\pm$ 0.20\% & 95.16\% $\pm$ 0.20\% \\
Verifiability \& Intent & 92.54\% $\pm$ 0.13\% & 92.96\% $\pm$ 0.09\% & 92.61\% $\pm$ 0.14\% & 92.73\% $\pm$ 0.20\% \\
Evidence Basis & 93.64\% $\pm$ 0.23\% & 93.67\% $\pm$ 0.09\% & 94.06\% $\pm$ 0.09\% & 94.25\% $\pm$ 0.26\% \\
Certainty & 94.30\% $\pm$ 0.09\% & 94.53\% $\pm$ 0.15\% & 94.55\% $\pm$ 0.07\% & 94.53\% $\pm$ 0.10\% \\
Temporality & 93.45\% $\pm$ 0.13\% & 93.50\% $\pm$ 0.22\% & 93.38\% $\pm$ 0.17\% & 93.67\% $\pm$ 0.24\% \\
Claim Focus & 91.30\% $\pm$ 0.25\% & 91.37\% $\pm$ 0.25\% & 91.05\% $\pm$ 0.03\% & 91.49\% $\pm$ 0.17\% \\
\hline
\end{tabular}
\label{tab:trt_robustness_full}
\end{table*}

{\subsection{Phase II Optimization: Heuristic Refinement across Contextual Windows}
\label{app:phase2_heuristics}}
To understand the interaction between heuristic injection and discourse context, we expanded our optimization analysis by evaluating the rule-based lexical heuristics across all six candidate context windows. In these ablation experiments, the exact same linguistic heuristics—governing structural transitions, explicit lexical anchors, and grammatical patterns—were appended to each respective context prompt. The results validate that while the introduction of rule-based heuristics uniformly elevates accuracy baselines across all configurations compared to raw context alone, their effectiveness is highly sensitive to the size of the discourse window. Most notably, Axis VI (Claim Focus) displays a critical architectural dependency: under every other contextual setting (such as isolated claims, full transcripts, or summaries), the accuracy of Claim Focus remains severely bottle-necked, dropping as low as 56.27\%. However, when these linguistic guidelines are paired precisely with the short-range Local +/- 2 window, the model successfully anchors the lexical cues. This dynamic drives the mean performance to its study-wide peak of 86.44\% accuracy ($\kappa = 0.76$).

\subsection{Multi-Temperature Robustness Analysis and Stability Breakdown}
\label{sec:robustness_appendix}
To evaluate the empirical stability and robustness of our structured discourse pipeline under stochastic decoding, we conducted a multi-temperature evaluation on the TRT domain dataset. We executed the complete pipeline across four decoding temperatures ($\tau \in \{0.0, 0.2, 0.5, 0.8\}$) with three independent trials per setting, totaling 12 execution runs.

\noindent\textbf{Analysis of Results and Stability:} 
As detailed in Table\mbox{~\ref{tab:trt_robustness_full}}, the standard deviations across all 12 runs are exceptionally low across all evaluation dimensions.
This high level of internal consistency demonstrates that our framework and rule-based heuristics produce stable, reproducible outputs rather than random generational artifacts, confirming the overall robustness of the structured prompting approach across varying decoding temperatures.

\noindent\textbf{Baseline Comparison:} 
When comparing these repeated-trial figures to the primary evaluation results in the main text, minor baseline performance variations are observable (e.g., a $3.70\%$ increase in Aspect Labeling, a $1.77\%$ increase in Stance Detection, and a $4.79\%$ increase in the 6-Axis Typology mean at Temperature 0.0, as shown in Table\mbox{~\ref{tab:baseline_comparison}}). These minor numerical differences stem from standard API execution environment updates and downstream library evolutions over time. Crucially, the relative performance tiers and core findings across tasks remain entirely consistent.

\subsection{Cross-Model Performance}
\label{sec:cross_model_full}
Table~\ref{tab:cross_model_full} reports complete results for all four models 
across all four domains under the identical B2-CoT pipeline configuration. 
While Gemini 2.5 Flash achieves the strongest overall performance, 
particularly on the 6-axis typology task, open-weight alternatives remain highly competitive. Specifically, Qwen 3.5 35B consistently approaches proprietary-model performance on aspect and stance classification across multiple thematic domains.  
In contrast, Llama 3.3 and Gpt-oss 20B exhibit more substantial performance degradation on the pragmatic typology task, highlighting a clear capability gap in handling high-dimensional rhetorical profiling.
As evidenced by these results, structured pragmatic discourse analysis imposes significantly greater reasoning and instruction-following demands than traditional thematic categorization, causing smaller or less specialized open-weight models to struggle under complex multi-dimensional inference settings.

\twocolumn[{
    \centering
    \small
    \setlength{\tabcolsep}{4.5pt}
    \captionof{table}{Cross-model performance comparison on Aspect labeling, Stance detection, and 6-Axis Typology classification under the identical B2-CoT pipeline configuration.}
    \label{tab:cross_model_full}
    \begin{tabular}{ll cc cc cc}
    \toprule
    & & \multicolumn{2}{c}{\textbf{Aspect Labeling}} 
      & \multicolumn{2}{c}{\textbf{Stance Detection}} 
      & \multicolumn{2}{c}{\textbf{6-Axis Typology (Mean)}} \\
    \cmidrule(lr){3-4}\cmidrule(lr){5-6}\cmidrule(lr){7-8}
    \textbf{Model} & \textbf{Domain} & Acc.\ (\%) & $\kappa$ & Acc.\ (\%) & $\kappa$ & Acc.\ (\%) & $\kappa$ \\
    \midrule
    \multirow{4}{*}{Gemini 2.5 Flash}
      & Ozempic  & \textbf{79.23} & \textbf{0.7663} & \textbf{91.55} & \textbf{0.8716} & \textbf{86.15} & \textbf{0.7583} \\
      & TRT      & \textbf{89.12} & \textbf{0.8813} & \textbf{92.15} & \textbf{0.8770} & \textbf{88.62} & \textbf{0.7676} \\
    & Collagen & \textbf{86.44} & \textbf{0.8475} & \textbf{85.59} & \textbf{0.7663} & \textbf{83.97} & \textbf{0.7249} \\
    & Fasting  & 77.64 & 0.7512 & 83.38 & 0.7326 & \textbf{82.23} & \textbf{0.5865} \\
    \midrule
    \multirow{4}{*}{Llama 3.3}
    & Ozempic  & 76.49 & 0.7364 & 83.86 & 0.7531 & 75.56 & 0.6095 \\
  & TRT      & 64.77 & 0.6147 & 82.56 & 0.7369 & 73.90 & 0.4880 \\
  & Collagen & 74.15 & 0.7105 & 75.85 & 0.5972 & 71.96 & 0.5244 \\
  & Fasting  & 67.37 & 0.6376 & 84.59 & 0.7536 & 69.79 & 0.4077 \\
    \midrule
    \multirow{4}{*}{Qwen 3.5 35B}
    & Ozempic  & 77.54 & 0.7474 & 85.26 & 0.7761 & 81.58 & 0.6983 \\
  & TRT      & 79.00 & 0.7708 & 83.30 & 0.7384 & 82.05 & 0.6221 \\
  & Collagen & 81.36 & 0.7896 & 78.39 & 0.6480 & 81.43 & 0.6815 \\
  & Fasting  & \textbf{78.25} & \textbf{0.7579} & 86.71 & 0.7869 & 80.72 & 0.5847 \\
    \midrule
    \multirow{4}{*}{Gpt-oss 20B}
    & Ozempic  & 78.60 & 0.7582 & 81.40 & 0.7171 & 73.04 & 0.5894 \\
  & TRT      & 71.17 & 0.6839 & 82.56 & 0.7319 & 75.33 & 0.5262 \\
  & Collagen & 71.61 & 0.6766 & 79.66 & 0.6629 & 75.85 & 0.5907 \\
  & Fasting  & 70.69 & 0.6716 & \textbf{87.31} & \textbf{0.7971} & 76.08 & 0.5181 \\
    \bottomrule
    \end{tabular}
    \vspace{1.5em} 
}]

\end{document}